\documentclass[runningheads]{llncs}

\usepackage{accv}

\usepackage{accvabbrv}

\usepackage{graphicx}
\usepackage{booktabs}

\usepackage[ruled,vlined]{algorithm2e}

\usepackage{tabularx}
\usepackage{booktabs}
\usepackage{booktabs}
\usepackage{array}
\usepackage{siunitx}

\usepackage[accsupp]{axessibility}  

\usepackage[pagebackref,breaklinks,colorlinks,citecolor=accvblue]{hyperref}

\usepackage{orcidlink}

\begin{document}

\title{EMPURPLE: A Free Lunch for Diffusion Distillation based on the Information Bottleneck} 

\titlerunning{Abbreviated paper title}

\author{Zilai Li\inst{1}\orcidlink{https://orcid.org/0009-0001-0804-6746} \and
Lujia Bai\inst{1}\orcidlink{https://www.researchgate.net/profile/Lujia-Bai} }

\authorrunning{Zilai Li, Lujia Bai}

\institute{Independent researcher, University of Nottingham, Nottingham, UK \and
 Ruhr University Bochum,  D-44801 Bochum, Ruhr, German
\email{lizilai2008@163.com}\\
\url{https://www.ruhr-uni-bochum.de/en}  }

\maketitle

\begin{abstract}

Diffusion models achieve impressive image-generation quality but remain expensive at inference time. Diffusion distillation reduces sampling steps, yet many distilled models, including SDXL-Lightning and distribution matching distillation methods, suffer from degraded Fr\'echet Inception Distance (FID). We analyze this phenomenon through a PAC-style generalization bound. Our analysis suggests that aggressive early-step redirection of the velocity field makes the distillation target harder to learn, enlarging the train-test gap. As a result, early-step output distributions differ between training and inference, causing distribution mismatch in the intermediate noisy latent used as next-step inputs. We empirically validate this mechanism by showing reduced diversity in both intermediate features and final outputs. To address this issue, we propose EMPURPLE, a simple training-free method that recycles intermediate latents sampled from the original model. EMPURPLE is model-agnostic and improves FID by 7\% to 20\% across DMD2, Hyper-SD, FlashSD, and SDXL-Lightning.

\end{abstract}

\section{Introduction}
Diffusion models (DMs)\cite{sohl2015deep} achieve state-of-the-art image generation quality (e.g., low FID\cite{heusel2017gans}) and are typically easier to train than GANs\cite{goodfellow2020generative,saatci2017bayesian}. Their main drawback is inference cost: sampling often requires many steps to solve a probability-flow ODE\cite{song2020score}.

Prior work accelerates sampling either by improving ODE solvers\cite{lu2022dpm,lu2025dpm,zhao2023unipc,zhou2024amed} or by distilling the multi-step sampler into a few-step model\cite{song2023consistency,yin2024dmd,yin2024improveddmd,lin2024sdxllightning,sauer2024adversarial}. While distillation can preserve perceptual quality, it often degrades FID and mode coverage.

We observe a consistent difference in few-step trajectories: ODE solvers produce a blurred latent early and remain deterministic, whereas distilled samplers (e.g., DMD2\cite{yin2024improveddmd}) tend to predict a detailed latent early and then inject noise. We argue this is wasteful: high-frequency details are hard to predict and are immediately corrupted by the random noise injection in the distillation algorithm.

We quantify this gap by comparing the covariance spectra of early blurred latents and the final detailed latents. Concretely, we compute covariance matrices and analyze their principal axes, as in PCA\cite{pearson1901liii}, where eigenvectors define orthogonal directions of variation and eigenvalues measure variance along them. A low effective rank\cite{roy2007effective} means the variance concentrates in few directions. For the predicted clean latent, the early-stage covariance has a much lower effective rank (\(\sim 70\times\) smaller) and a smaller eigenvalue sum (total variance), indicating a lower-entropy, easier prediction problem. Distillation, which predicts detailed latents, trains to match a harder target than the original diffusion model, letting the output of the distillation model in inference not be good enough. While the original diffusion model, which studies an easier task, can ensure the next step input is in-distribution in inference.

\begin{figure}[ht]
  \vskip 0.2in
  \begin{center}
    \centerline{\includegraphics[width=\columnwidth]{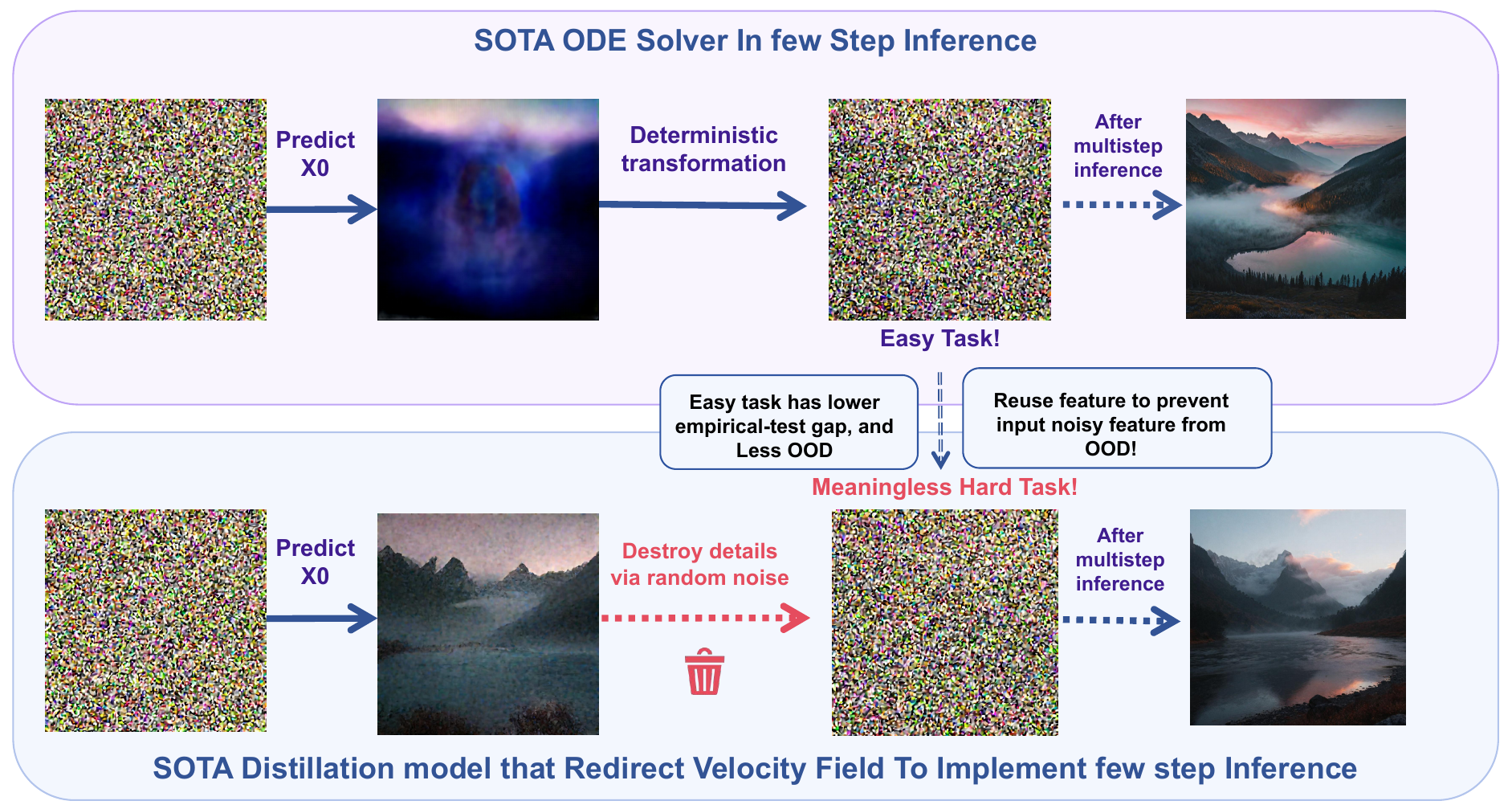}}
  \caption{Overview of EMPURPLE. The original diffusion model produces a blurred latent early in sampling, whereas a distilled model often produces a detailed latent and then adds noise. This aggressive redirection can increase the generalization gap and induce a train--test distribution mismatch in intermediate latent. EMPURPLE mitigates the mismatch by reusing cached intermediate latent sampled from the original model.}
  \label{Empurple-algorithm}
  \end{center}
\end{figure}

Based on this analysis, we propose \textbf{EMPURPLE} (Enhance Model Painting Using Recycled and Proper Latents is Easy), a training-free method (\cref{Empurple-algorithm}). EMPURPLE caches 10{,}000 intermediate noisy latents from the original diffusion model and reuses them to initialize distilled sampling, keeping early-step inputs in-distribution. Experiments show that DDIM inversion with random-prompt guidance can encode early noisy latents to Gaussian noises largely remaining in the Gaussian typical set; under the standard assumption that when low-CFG is low,  DDIM inverse/forward approximately match\cite{mokady2023null}, reusing latents approximates running the original model in the early stage, where predicting blurred structure is easier and helps prevent OOD inputs downstream.

\section{Background}
\subsection{Probability Flow ODE and Diffusion Distillation}

Diffusion models in computer vision typically aim to transport Gaussian noise $x_0\sim p_0$ into an image sample $x_1\sim p_1$. During training, rectified flow \cite{liu2022rectified} defines a non-causal transport trajectory, shown in \cref{eq:extension_x_t}.
\begin{equation}
   X_t = \alpha_t X_1+ \beta_t X_0,~ u^{cond}(x_t,x_1,t) = \frac{-\dot{\beta}_t\alpha_tx_1+\dot{\beta}_tx_t+\dot{\alpha}_t\beta_tx_1}{\beta_t}
  \label{eq:extension_x_t}
\end{equation}
Here, $\alpha_t$ and $\beta_t$ are researcher-chosen schedules; $\dot{\alpha}_t$ and $\dot{\beta}_t$ denote their derivatives with respect to $t$; and $u^{\mathrm{cond}}(x_t,x_1,t)$ describes the non-causal velocity field that transports $X_t$ to $X_1$ during training. Different choices of $\alpha_t$ and $\beta_t$ recover linear rectified flow \cite{liu2022rectified} or a variance-preserving ODE \cite{karras2022elucidating,song2020score}.


Since \cref{eq:extension_x_t} is none-causal, in the inference, 
based on the continuity equation \cite{bertrand2025closed,liu2022rectified}, we could only use a neural network approximating the \cref{eq:gt_v_f}. 
\begin{equation}
   u^{*}(x_t,t) = \mathbb{E}_{x_1|x_t}[u^{cond}(x_t,x_1,t)]
  \label{eq:gt_v_f}
\end{equation}

That means the ground truth velocity field point to the average of images that can cause the current noisy $x_t$ by adding Gaussian noise to them. And we use a neural network $u_\theta(x_t,t)$ parameterized by $\theta$ to study it. According to the continuity equation\cite{liu2022rectified}, this ground truth velocity field can transport an arbitrary $p_0(Z_0)$ to an arbitrary $p_1(Z_1)$ with an arbitrary relation $p(Z_1|Z_0)$, and the image generation process is just a special case; image editing can also be described by this ODE\cite{liu2022rectified}. 
In the distillation, the change of the $p(Z_1|Z_0)$ also modifies the expectation, letting the transportation trajectory straighter\cite{liu2022rectified}, which redirects the velocity field and lets it point to a less blurred output.

\subsection{PAC Bound, Information theory and the key source of the generalization of the diffusion model}



previous research \cite{li2026fschedule} reveals the relationship between the probably approximately correct (PAC) \cite{hoeffding1963probability} bound and the diversity problem, which claims that an algorithm's generalization ability, relating to the expected error on the unseen data, is also affect the diversity of output images. 

Explicitly, consider the well-known PAC deduction from Hoeffding's inequality, for Gaussian noise $X\sim \mathcal{X}$ and corresponding generative result $Y$ output by an model with better FID performance, like another distillation model with better optimization target, we have \cref{eq:HoeffdingInequality} for the distillation model $f^d$ trained in dataset $d$ with validation set $\{x_i,y_i\}_{i=1}^m\sim S$. The increase in the loss will change the output probability distribution.
\begin{equation}
   \Delta(D) = E_{X, Y}[\mathcal{L}(f^d(X) ,Y)] - \frac{1}{m} \Sigma_{i=1}^m\mathcal{L}(f^d(x_i),y_i)\leq\sqrt{\frac{\log{|\mathcal{H}|+\log{\frac{2}{\sigma}}}}{2o}},\label{eq:HoeffdingInequality}
\end{equation}
the \cref{eq:HoeffdingInequality} correct with probability approaching 1. Explicitly, a hypothesis is a function a neural network can express based on the observe data, and the $\mathcal{H}$ is the hypothesis space that contains all of the function a model can express, $o$ is a given constant for each hypothesis space, $x$ and $y$ are samples of Gaussian noise and images,  $f^d(X)$ denotes a particular neural network trained in dataset $s$.  And the $\sigma$ denotes the probability $1-\sigma$ that the inequality is correct.

\section{Approach}

\subsection{Analysis on the Mechanism Of How PAC Bound Affects Diffusion Distillation}

The diffusion model includes multiple neural network evaluations. To understand the PAC bound in \cref{eq:HoeffdingInequality}  affecting which part of the distillation algorithm, we should first examine the mechanism of distillation training. 

There is a trade-off in the diffusion distillation: 

1. To decrease the inference step, researchers tend to increase the moving distance on each step, which requires the trajectory of the causal transportation given by the neural network to be as straight as possible \cite{liu2022rectified}.

2. To train the diffusion distillation effectively, researchers tend to add noise on the training image to get noisy latent $Z_t$ and let the neural network output a redirected velocity field by inputting a noisy latent $Z_t$ \cite{chadebec2025flash,yin2024dmd,yin2024improveddmd,luo2311lcm,lin2024sdxllightning} .


The second point highlights a key feature of distillation training: the distillation model must take an input of noisy latent whose distribution matches the inputs seen by the original diffusion model. By contrast, the first point suggests a different requirement: the distillation model should transport $p_0(Z_0)$ to $p_1(Z_1)$ along a trajectory that deviates from the original model's, and the intermediate distribution $p_t(Z_t)$ may different from the distribution given by the original trajectory. To cope with this problem, most prevalent algorithms \cite{yin2024dmd,yin2024improveddmd,lin2024sdxllightning,luo2311lcm} require the distillation model to directly output an image $Z_1\sim p_1(Z_1)$ at each point of the trajectory using one-step inference, and then add noise to the output to further improve the one-step prediction. If successful, adding noise to $Z_1\sim p_1(Z_1)$ ensures that the intermediate noisy latents have identical distributions during inference and training. However, if $Z_1$ can be generated in one step, multi-step inference is unnecessary; accordingly, DMD2 \cite{yin2024improveddmd} adds noise to the image output by the neural network and uses it as the input to the next training step.


However, in the latter part of this section, we also show that when the distillation algorithm requires a neural network to change the blurred output given by \cref{eq:gt_v_f} to a detailed latent with high-frequency components, the training target becomes more difficult. As a consequence, the gap between empirical loss and expected loss increases, and then the input probability distribution in the next step will be different between training and inference.  Hence, DMD2 would still confront the mismatch of the input probability distribution in training.  The output of the distillation model \cite{yin2024dmd,yin2024improveddmd,chadebec2025flash,luo2311lcm} is perturbed by random noise, and it mainly contributes to the low-frequency component of the next step input.
Additionally, in \cref{sec:low_freq_comp}, we add noise to the final generative result and use the DDIM inverse to encode it, and find that the final result loses the diversity on low-frequency components.
To explain these results, we provide theoretical and empirical evidence in the following section. First, we show that predicting a detailed image is a more challenging learning task. Second, we show that such prediction changes the distribution of the next-step input, creating a mismatch between training and inference. Our experiments further demonstrate that the inference-time distribution loses diversity under this mismatch.

\subsubsection{Predicting the detailed image is a more difficult task.}

The basic intuition is that the detailed image contains more information. To prove this, we first compute the covariance matrix of 10,000 examples for both the blurred image and the detailed output, as in the PCA algorithm \cite{pearson1901liii}. Based on the covariance matrix, we calculate the effective rank \cite{roy2007effective}, which provides the number of axes given by the eigenvectors of the covariance matrix that have sufficiently high variance, and compute the sum of the eigenvalues (the variance) of the covariance matrix.  We find that both the effective rank and the eigenvalues of the blurred image are clearly lower than those of the detail image. This implies that the entropy, which is mainly determined by the variance, is higher.
 Meanwhile, for a random variable, the volume of the typical set is $2^{H(X)}$ \cite{cover1999elements}.  And the data of the blurred image lies in the low-dimensional and low-variance manifold, which also means the volume is smaller. The detailed result of the covariance statistic is in \cref{sec:vol_comp}.  In computer science, we don't handle infinite precision input; the above observation means that when the input of two tasks in the training belongs to the same probability distribution, the task with fewer possible outputs has a lower upper bound on the number of functions it can express.

\subsubsection{The shift of the probability distribution in inference and training.} 

To strictly prove the shift of the probability distribution when the training target change, we should first provide a mathematical definition:
For a neural network $f^d$ with a given optimization method and a particular setting of the hyperparameters of the optimizer, the function it can express after the training with $i$ times iteration constitutes the hypothesis space $\mathcal{H}$, and we can define a special function it can express as a hypothesis $h$ by providing a particular training dataset $d\sim D^m$ . And the \cref{eq:HoeffdingInequality} can be write as the gap between the empirical loss $\hat{R}(h)$ and expected loss $R(h)$.  When given a special training dataset $d$ and a input $X$, for a deterministic neural network, the $i$th gradient descent (GD) \cite{ruder2016overview} optimizaiton would provide a fixed value as output, and the $\mathcal{L}$ in \cref{eq:HoeffdingInequality} could be a simple L2 loss, while the possible output $Y$ with given $X$ in $i$th stochastic gradient descent (SGD) \cite{ketkar2017stochastic} optimization is a random variables, and the typical loss is the log likelihood. 

However, the agnostic on the shape of the distribution prevents us from directly utilizing the log likelihood; hence, we utilize the following loss in the paper to help us prove the shift of the distribution.  
\begin{equation}
\mathcal{L}(f^d(X),Y)=\mathbb{E}_{f\sim \mathcal{F}}[(f(X)-Y)^2],
  \label{eq:new metrics}
\end{equation}
 in which $\mathcal{F}$ denotes all of the possible result functions in the $i$th iteration on the given SGD optimization with the given training dataset $d$.  

Then, the expected loss in \cref{eq:new metrics}, based on the bias-variance trade-off, can be decomposed into two parts. The precise proof is in \cref{app:vb-tradeoff} of the appendix. 

\begin{proposition}
\label{prop:collapse}
$ \mathbb{E}_{X\sim \mathcal{X},Y=T(X)}[\mathcal{L}(f^d(X),Y)] = \mathbb{E}_{X\sim \mathcal{X},Y=T(X)}[(f'(X)-Y)^2]\\ + \mathbb{E}_{X\sim \mathcal{X},f\sim \mathcal{F}}[(f(X)-f'(X))^2] $
, where $\hat{f}^d(x)=E_{f\sim\mathcal{F}}[f(x)],f'(x)=E_{X\sim \mathcal{X}}[\hat{f}^d(x)]$
  \label{eq:decomposeLose-New}
\end{proposition}
, where $T(X)$ is a function provide by a better model. The proposition implies that the expected loss decomposes into (i) the difference between the mean output and the mean ground-truth image and (ii) the output variance. Thus, a loss gap between test and training will shift the output distribution, and propagate to later inputs. Even if distillation training does not add noise to real images to get the next step input, but uses input obtained by adding noise to the training model's output, the problem of input distribution mismatch still exists.

In the following part, we would prove that the original diffusion model's target (a blurred image) is easier to learn, and therefore yields a smaller empirical-test gap than distillation, then the train-inference distribution mismatch should be less severe.

\begin{figure}[ht]
  \vskip 0.2in
  \begin{center}
    \centerline{\includegraphics[width=\columnwidth]{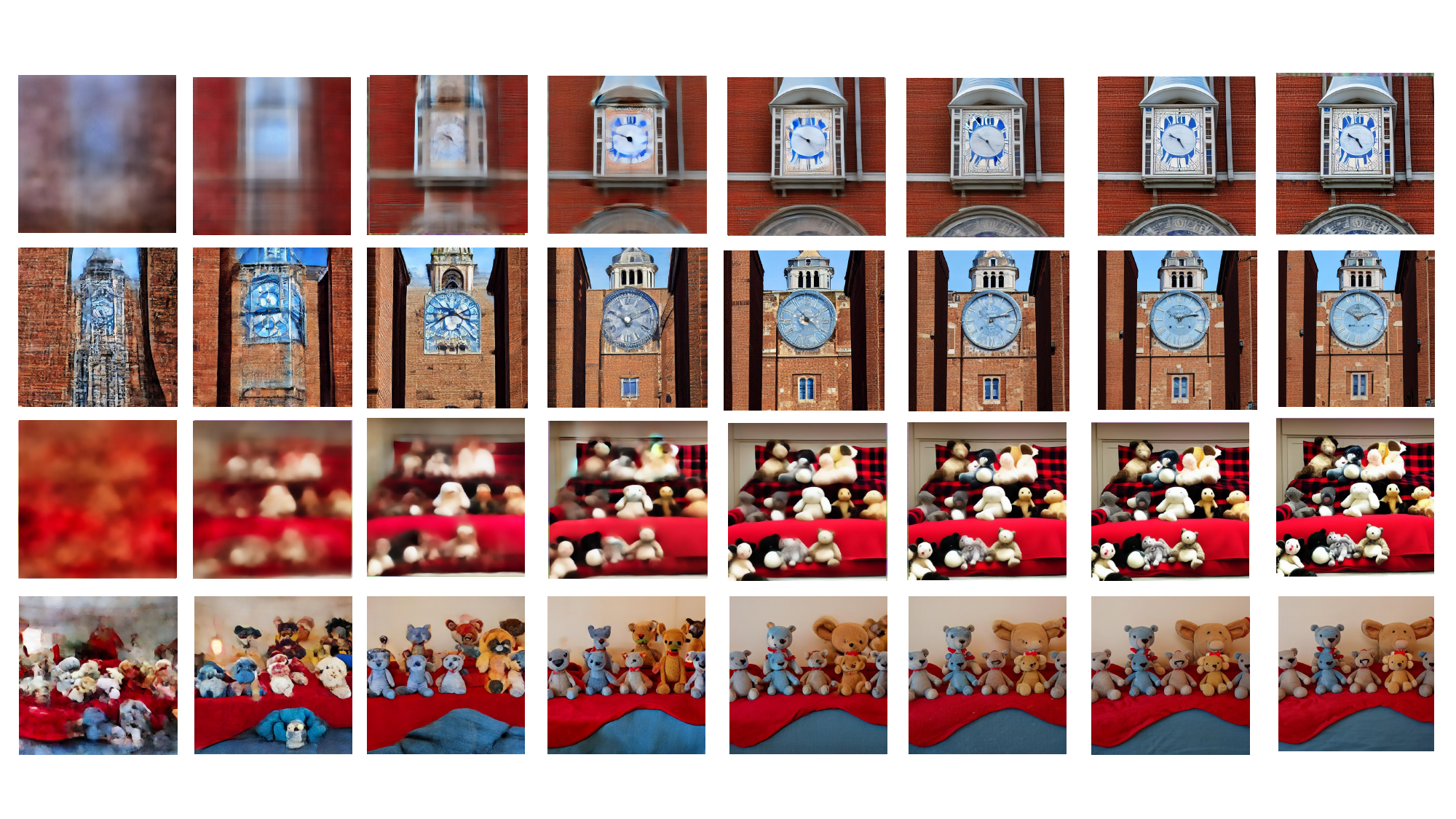}}
  \caption{Visualize the middle output in the inference process of the original diffusion model.  The first and third lines are the output from the original diffusion model, while the second and fourth line are the output from the Flash SD.}
  \label{middleFeature}
  \end{center}
\end{figure}

Explicitly, for a diffusion model, the inference process provides a Markov chain $Z_0^1$-$Z_{t_1}^1$-$Z_{t_{2}}^1$-...--$Z_1^1$, in which $Z_t^1$ is the noisy input in the inference. We only discuss the mechanism between $Z_{t_i}^1$-$Z_{t_{i+1}}^1$ in the early inference stage.  Explicitly, in the distillation model, after the training, the velocity field described by \cref{eq:gt_v_f} will point to different clean latent $\hat{Z}_1(Z_{t_i})$. For example, the DMD algorithm tend to let the velocity field points to the final inference result, while the progressive distillation redirect to the velocity field output by input $Z_{t_{i+2}}$ to halve the inference step.   
And for the original diffusion model, the velocity field point to a blur centroid, whose symbol is $\hat{Z}_1'(Z_{t_i})$. We visualize the inference process in the \cref{middleFeature}

Then, the mutual information between $\hat{Z}_1(Z_{t_i})$ and $Z_{t_i}$ in the distillation model is 
\begin{equation}
    I(\hat{Z}_1(Z_{t_i}),Z_{t_i})= H(\hat{Z}_1(Z_{t_i}))-H(\hat{Z}_1(Z_{t_i})|Z_{t_i})\leq H(\hat{Z}_1(Z_{t_i}))
    \label{distillation_MI}
\end{equation}
, and for the original diffusion model, who predict the blur centroid of possible images $\hat{Z}_1'(Z_{t_i})$, it's:
\begin{equation}
    I(\hat{Z}_1'(Z_{t_i}),Z_{t_i}')= H(\hat{Z}_1'(Z_{t_i}'))-H(\hat{Z}_1'(Z_{t_i}')|Z_{t_i}') \leq H(\hat{Z}_1'(Z_{t_i}'))
    \label{org_MI}
\end{equation}
In both \cref{org_MI} and \cref{distillation_MI}, the SGD optimization process gives the conditional entropy term.  Those equations mean the entropy of the  $\hat{Z}_1'(Z_{t_i})$ and $\hat{Z}_1(Z_{t_i})$ is the upper bound of the mutual information. 



And we have \cref{eq:gap-New} for the distribution shift. The precise proof is in the \cref{app:pac-b-mi} of the appendix.  
\begin{proposition}\label{prop:collapse}
$\nonumber\Delta(s) = \mathbb{E}_{X\sim \mathcal{X},Y=T(X)}[(f'(X)-Y)^2]  -\mathbb{E}_{X\sim S,Y=T(X)}[(f''(X)-Y)^2] + \mathbb{E}_{X\sim \mathcal{X},f\sim \mathcal{F}}[(f(X)-f'(X))^2] - \mathbb{E}_{X\sim S,f\sim \mathcal{F}}[(f(X)-f''(X))^2] \\ \leq \sqrt{\frac{{I(X;Y)}2^{I(X;Y)}+\log{\frac{2}{\sigma}}}{2o}}$
, where $\hat{f}^d(x)=E_{f\sim\mathcal{F}}[f(x)],f'(x)=E_{X\sim \mathcal{X}}[\hat{f}^d(x)]\\,f''(x)=E_{X\sim S}[\hat{f}^d(x)]$
  \label{eq:gap-New}
\end{proposition}
The proposition means that the less the mutual information between output and input, the less the test-train gap. If the training target forces the output to contain more mutual information, then the mismatch of the input probability distribution in variance and mean would also be increase.  Based on this, we reuse the middle noisy latent provided by the early inference statge of the original diffusion model to handle the mismatch in the distillation inference.

\subsection{EMPURPLE: reuse feature to cope with the mismatch of probability distribution}
 Based on the analysis of the previous two points, we propose EMPURPLE to circumvent the problem of predicting detailed images, improve the diversity of the generated images, and save computation.
In our algorithm, we cache $Z_{t}$ from the original diffusion model during early inference and reuse it during distilled inference. This ensures that the distillation model’s intermediate noisy latent variables follow the same distribution as in the original inference process.


Explicitly, the cache noisy feature is given via a previous inference of the original diffusion model.  If in a counterfactual situation, in which we choose another prompt and a different Gaussian noise, we still can get this middle noisy latent $Z_t$ in new inference, then, reuse $Z_t$ is equal to do the inference with using the original diffusion model in the early inference stage.
To prove this, we will use DDIM inverse on the cache noisy feature via the prompt guidance from the other random caption, and then calculate the probability of the encoded noise, to see whether it's in the typical set \cite{cover1999elements} of the standard Gaussian distribution. 

And experiment prove that, when we choose a random caption to guide the encoding, 99.92\% of encode noise is from the typical set of the standard Gaussian distribution, hence, if the assumption in the previous research \cite{mokady2023null} is correct, that the DDIM inverse and forward is almost identical in low classifier-free guidance (CFG) \cite{ho2022classifier} situation, reusing the cached feature is equal to use a new noise in the typical set to do a new inference via another prompt as a gudiance.  

Based on this, we proposed a training-free algorithm in \cref{tab:algorithm1} and \cref{tab:algorithm2} to enhance the inference.  In the \cref{tab:algorithm2}, we could choose not to randomly pick up a cache noisy feature, but calculate the cosine similarity between the cached prompt embed and the current prompt embed to get the corresponding feature.

\begin{algorithm}
  \SetAlgoLined
  \KwData{
  special timestep: $t$
  , $m$ prompt embeds from arbitrary training dataset: $C_m=\{{c_i}\}_{i=1}^m$
  , original diffusion model: $u_\theta'$, arbitrary PF ODE Solver start at 0 and terminate on $t$: $\Phi'(u_\theta',z,0)$,}
  \KwResult{ a list of cached features $Z_m=\{\hat{z}_t^i\}_{i=1}^m$
  }
  \LinesNumbered
  
  $Z_m$ =[] 

  \While{$c_i$ in $C_m$}{
      $\epsilon\sim\mathcal{N}(0,I)$; 
      
      $\hat{z}_t^i = \Phi'(u_\theta,\epsilon,0)$;
       
      $Z_m.append(\hat{z}_t^i)$;
  } 
  
  \caption{Empurple: Data prepare process}
  \label{tab:algorithm1}

\end{algorithm}

\begin{algorithm}
  \SetAlgoLined
  \KwData{
  special timestep: $t$, list of cache feature $Z_m=\{\hat{z}_t^i\}_{i=1}^m$, $m$ prompt embeds from arbitrary training dataset: $C_m=\{{c_i}\}_{i=1}^m$
  , a series of prompt embeds that guide the current inference: $C'_n=\{{c_i'}\}_{i=1}^n$
  , arbitrary pretrained distillation model: $u_\theta$
  , arbitrary PF ODE/SDE Solver start on time t: $\Phi(u_\theta,z,t)$, bool value determined whether randomly fetch cached feature: $B$}
  \KwResult{final result image $z_1$
  }
  \LinesNumbered

  \While{$c_i'$ in $C_n'$}{ 
\If{not B}{%
  \While{$c^i$ in $C_m$}{%
    list.append(\text{cosine\_similarity}($c^i, c_i'$))
  }%
  
  label $\leftarrow \operatorname{softmax}(\text{list})$\\
  idx $\leftarrow \operatorname{multinomial}(\text{label})$
}
\Else{
  idx $\sim \mathcal{U}(0, m)$
}
    
   $z_1 = \Phi(u_\theta,\hat{z}_t^{idx},t)$ 
   }
  \caption{Empurple: Inference process}
  \label{tab:algorithm2}
\end{algorithm}

\section{Experiments}

\subsection{Compare the Volume of Different Features' typical set}
\label{sec:vol_comp}
To validate our intuition (\cref{sec:vol_comp}) that the early-step blurred latents carry less information---i.e., have lower entropy and therefore a smaller typical-set volume---than the final-step detailed latents used as the distillation target, we compare the volume statistics of these two feature distributions. Concretely, we run a 1-step inference and collect 10,000 blurred latents at timestep 999 with $CFG=5.5$. We then compute the empirical covariance matrix and report (i) its effective rank, (ii) the mean eigenvalue, and (iii) the sum of eigenvalues over the effective-rank subspace. We compute the same quantities for the latents encoded from the final generated images. The blurred features show an effective rank about $70\times$ smaller and a substantially smaller eigenvalue mass, indicating both fewer high-variance directions and lower overall variance. Full results are reported in \cref{tab:pixel-mean-comparison}.

\begin{table}[t]
\centering
\small
\setlength{\tabcolsep}{5pt}
\renewcommand{\arraystretch}{1.15}
\caption{Comparison of statistic characteristic of the covariance matrix of the latent variables output in the first and the final step.}
\label{tab:pixel-mean-comparison}
\begin{tabularx}{\linewidth}{
  >{\raggedright\arraybackslash}X
  c
  c
  c
}
\toprule
{\shortstack{output\\stage}} & {\shortstack{Effective\\rank}}  & {\shortstack{sum of eigenvalues\\in the effective rank}} & {\shortstack{means of\\eigenvalues}} \\
\midrule
Fist step & 36.363132 & 3379  & 0.385991 \\
Final step & 2268.534180 & 5636 & 0.750580 \\
\bottomrule
\end{tabularx}
\end{table}

\subsection{DDIM Inverse Noise Comparison}
\label{sec:nois_comp}


We use the F-scheduler \cite{li2026fschedule} as an 8-step solver for the PF ODE, sampling 10{,}000 images guided by captions from the COCO 2014 training set \cite{lin2014coco}. During inference, we cache first-step output lantents together with the corresponding prompt embedding for later reuse. All 8-step scheduler hyperparameters follow the official configuration. Because the cached noisy feature from the original model is reused during validation, we also cache the first step output from the inference applying the distilled model to the validation-set caption, enabling a direct comparison of noisy-feature quality.

We use random captions from the COCO 2014 validation set to guide a 4-step DDIM inversion, which encodes the noisy features produced in the first inference step. Empirically, 99.92\% of the recovered Gaussian noise from the original diffusion model lies in the Gaussian typical set. Based on the assumption in previous research\cite{mokady2023null}, which claims the DDIM inverse and forward have almost identical trajectories when the CFG is low, experiments support our claim that the cached early-stage noisy features can be reproduced in a new inference run by pairing a different random prompt with a Gaussian noise sample drawn from the same typical set.  The noisy feature of the SOTA distillation model, Flash SD, also recovers Gaussian noises that reach a high proportion in the typical set.  But its L1 distance to the original point is obviously smaller than the noisy feature from the original SD, which means the original diffusion model can use part of the Gaussian distribution to generate those noisy features when $CFG=1.0$.

Explicitly, based on the Asymptotic Equipartition Property (AEP) \cite{deco1996information}, we define the typical set $A^{n}_\epsilon$ for an i.i.d. Gaussian sequence $(x_1,\dots,x_n)\in\mathcal{X}^n$ with $n=64\cdot64\cdot4$ and $\epsilon=0.02$ (in nats). By definition, all noise whose probability density satisfies the premise of the \cref{typicalSet} is in the typical set.
\begin{equation}
  e^{-n(H(X)+\epsilon)} \le P(x_1,\dots,x_n) \le e^{-n(H(X)-\epsilon)}.
  \label{typicalSet}
\end{equation}
The noise encoded by DDIM inversion corresponds to one such typical sample.

In addition to random-caption guidance, we also encode noisy features from the inference of the original SD using prompts that are highly similar to the cached prompt embedding. Concretely, for each encoding process, we compute cosine similarities to the cached embeddings, apply a softmax to obtain a sampling distribution, and then sample a prompt embedding accordingly, excluding the highest-probability candidate each time. Results are reported in \cref{tab:typical-set-comparison}.


\begin{table}[th]
\centering
\small
\setlength{\tabcolsep}{5pt}
\renewcommand{\arraystretch}{1.15}
\caption{Comparison of mean pixel values under different settings.}
\label{tab:typical-set-comparison}
\begin{tabularx}{0.8\linewidth}{
  >{\raggedright\arraybackslash}X
  c
  c
  c
  c
}
\toprule
{\shortstack{model create\\Noisy latent}} & {\shortstack{compare\\cosine similarity}} & {\shortstack{Pixel\\mean}} & {\shortstack{$\ell_1$\\norm}} & {\shortstack{Typical-set\\ratio}} \\
\midrule
SD & No & 0.001 &   13052.34  & 99.92\% \\
Flash SD & No & 0.002 &  13016.97 & 99.91\% \\
LCM & No & 0.004 &  12925.31  & 96.37\% \\
\midrule
SD & Yes & 0.001 & 13053.73  & 99.87\%  \\
\bottomrule
\end{tabularx}
\end{table}

\subsection{Validate the lack of the low frequency components in the generative result of the distillation model}
\label{sec:low_freq_comp}
To test whether distilled models miss low-frequency content, we  (i) VAE-encode images from distilled models (LCM, FlashSD; 4 steps, official settings) and from Stable Diffusion 1.5 (ODE; 8-step F-scheduler \cite{li2026fschedule} with UniPC and CFG=5.5), (ii) add noise at level 800/1000, and (iii) run a 4-step DDIM inversion \cite{mokady2023null} to recover the corresponding Gaussian noise. We use COCO 2014 val captions and guidance scale 5.5 (FlashSD without CFG) to generate 10,000 samples, and report the encoded noise norm/mean and typical-set proportion (\cref{tab:middle-noise-shift}), plus FID and CLIP. The inverted noises from distillation model deviate from zero-mean and have smaller-than-expected L1 norms, suggesting that if the assumption in \cite{mokady2023null} correct, the DDIM forward/inverse trajectory is identical when $CFG{=}1$, then DDIM forward would then use that subset to match the distilled-model features.
We also sample 10000 random Gaussian noise 10000 times, the max value and min value of the average L1 norm is 13069 and 13075, which is obviously different from the encode Gaussian noise. 


\begin{table}[t]
\centering
\small
\setlength{\tabcolsep}{4pt}
\renewcommand{\arraystretch}{1.15}
\caption{Validation of the distribution shift in intermediate noisy features.}
\label{tab:middle-noise-shift}
\begin{tabularx}{\linewidth}{
  >{\raggedright\arraybackslash}X
  S[table-format = -1.8]
  S[table-format = 5.2]
  S[table-format = 1.4]
  S[table-format = 2.2]
  S[table-format = 2.2]
}
\toprule
Model & {\shortstack{Pixel\\mean}} & {\shortstack{$\ell_1$\\norm}} & {\shortstack{Typical-set\\ratio}} & {FID} & {\shortstack{CLIP\\score}} \\
\midrule
LCM & 0.00240000 & 13001.07 & 0.9972 & 23.61 & 30.72 \\
Flash SD & -0.00009928 & 13036.34 & 0.9997 & 16.18 & 30.64 \\
SD & -0.00004298 & 13060.90 & 0.9997 & 15.32 & 31.13 \\
Random Gaussian Noise & 0.00008350 & 13073.12 & 0.9995
& \multicolumn{1}{c}{--}
& \multicolumn{1}{c}{--} \\
\bottomrule
\end{tabularx}
\end{table}

\subsection{Empurple ablation study}

In \cref{tab:ablation}, we use different distillation model to generate images via the guidance of 10k captions from the COCO 2014 val dataset.  Explicitly, those images we generated have a 1024x1024 resolution.  And for those models utilizing EMPURPLE, they just do the 4-step inference, reuse the cached feature mentioned in \cref{sec:nois_comp}.  And for other distillation models, since Empurple is equivalent to a 5-step inference, we do the 5-step inference, using the same time schedule as the EMPURPLE algorithm, and prove the increase of the inference step and time schedule could not improve the FID.
\begin{table}[th]
    \centering
    \setlength{\tabcolsep}{8pt}
    \renewcommand{\arraystretch}{1.2}
    \caption{FID and CLIP scores on the COCO 2014 validation set at $1024\times1024$ resolution. For rows without EMPURPLE, a 5-step inference schedule is used, matched in wall-clock time to the EMPURPLE setting. Lower is better for FID and higher is better for CLIP.}
    \label{tab:ablation}
    \begin{tabular}{lcccc}
        \toprule
        \textbf{Distillation model} & \textbf{EMPURPLE} & \textbf{Cosine similarity} & \textbf{FID} $\downarrow$ & \textbf{CLIP} $\uparrow$ \\
        \midrule
        SDXL-Lightning & No  & --  & 22.62 & 31.13 \\
        SDXL-Lightning & Yes & Yes & 18.21 & 29.55 \\
        SDXL-Lightning & Yes & No  & 18.55 & 29.49 \\
        \midrule
        Hyper SDXL     & No  & --  & 25.83 & 31.48 \\
        Hyper SDXL     & Yes & Yes & 21.95 & 30.20 \\
        Hyper SDXL     & Yes & No  & 20.85 & 30.04 \\
        \midrule
        DMD2           & No  & --  & 20.39 & 31.53 \\
        DMD2           & Yes & Yes & 16.05 & 30.87 \\
        DMD2           & Yes & No  & 15.84 & 30.81 \\
        \midrule
        Flash SDXL     & No  & --  & 21.74 & 31.19 \\
        Flash SDXL     & Yes & Yes & 19.10 & 29.49 \\
        Flash SDXL     & Yes & No  & 18.95 & 29.52 \\
        \bottomrule
    \end{tabular}
\end{table}
\subsection{Empurple performance}

In this paper, we validate the EMPURPLE algorithm for both 512x512 and 1024x1024 image generation via the metrics FID \cite{yu2021frechet} and CLIP Score\cite{hessel2021clipscore}.  When it comes to generating 512x512 images, we first use the caption from the COCO 2014 training dataset, Stable Diffusion, and the F-schedule solver to obtain the noisy latent features and prompt embeddings at 857/1000 noisy level.  Then, we apply EMPURPLE in \cref{tab:algorithm2} to the inference guide for the caption samples from the COCO 2014 and 2017 validation sets.  The number of caption samples from the COCO 2014 is 10k, while in 2017 it is 5k. We apply EMPURPLE in Flash SD\cite{chadebec2025flash}, LCM\cite{luo2311lcm}, and stable XL-turbo\cite{sauer2024adversarial}. 
Specifically, for SDXL-turbo, we only measure the generative result that uses 1-step inference with a randomly chosen latent embedding to augment the inference.
For all of the distillation models we tested, the only model that can set the guidance scale is the LCM model, for which we set its guidance scale to 8.0. The result is in \cref{tab:fid_clip_coco_sd}.

For 1024x1024 resolution generation, we use captions from the COCO 2014 training dataset, Stable Diffusion XL, and an F-scheduler to obtain 10k noisy feature and prompt embeddings at 852 timesteps, which is one of the inference steps.  The 1024x1024 F-scheduler, as the setting in the paper \cite{li2026fschedule}, use analytical first step (AFS). Then, we apply EMPURPLE to the inference guide for the caption samples from the COCO 2014 and 2017 validation sets.  The number of caption samples from the COCO 2014 is 10k, while in 2017 it is 5k. We apply EMPURPLE in Flash SDXL\cite{chadebec2025flash}, DMD2\cite{yin2024improveddmd}, and SDXL-lightning\cite{lin2024sdxllightning}, Hyper SDXL\cite{ren2024hyper}.  
Explicitly, for DMD2, Flash SDXL, we use LCM solver.  For SDXL-lightning and Hyper SD, we use DDIM solver.  The result is in \cref{tab:fid_clip_coco_sdxl}.

\begin{table}[ht]
    \centering
    \setlength{\tabcolsep}{8pt}
    \renewcommand{\arraystretch}{1.2}
    \caption{FID and CLIP scores for SD distillation models on COCO validation sets at $512\times512$ resolution. Lower is better for FID and higher is better for CLIP.}
    \label{tab:fid_clip_coco_sd}
    \begin{tabular}{lcccc}
        \toprule
        \textbf{Distillation model} & \textbf{EMPURPLE} & \textbf{Cosine similarity} & \textbf{FID} $\downarrow$ & \textbf{CLIP} $\uparrow$ \\
        \midrule
        \multicolumn{5}{c}{\textbf{COCO 2014 ($512\times512$)}} \\
        \midrule
        LCM        & No  & --  & 23.61 & 30.72 \\
        LCM        & Yes & Yes & 21.56 & 30.46 \\
        LCM        & Yes & No  & 21.35 & 30.45 \\
        Flash SD   & No  & --  & 16.18 & 30.64 \\
        Flash SD   & Yes & Yes & 14.66 & 29.78 \\
        Flash SD   & Yes & No  & 14.54 & 29.78 \\
        \midrule
        \multicolumn{5}{c}{\textbf{COCO 2017 ($512\times512$)}} \\
        \midrule
        LCM         & No  & --  & 30.29 & 30.60 \\
        LCM         & Yes & Yes & 27.75 & 30.42 \\
        LCM         & Yes & No  & 27.67 & 30.44 \\
        Flash SD    & No  & --  & 22.59 & 30.63 \\
        Flash SD    & Yes & Yes & 20.79 & 29.78 \\
        Flash SD    & Yes & No  & 21.10 & 29.77 \\
        \bottomrule
    \end{tabular}
\end{table}

\begin{table}[ht]
    \centering
    \setlength{\tabcolsep}{8pt}
    \renewcommand{\arraystretch}{1.2}
    \caption{FID and CLIP scores for SDXL distillation models on COCO validation sets at $512\times512$ and $1024\times1024$ resolutions. Lower is better for FID and higher is better for CLIP.}
    \label{tab:fid_clip_coco_sdxl}
    \begin{tabular}{lcccc}
        \toprule
        \textbf{Distillation model} & \textbf{EMPURPLE} & \textbf{Cosine similarity} & \textbf{FID} $\downarrow$ & \textbf{CLIP} $\uparrow$ \\
        \midrule
        \multicolumn{5}{c}{\textbf{COCO 2014 ($512\times512$)}} \\
        \midrule
        SDXL-turbo & No  & --  & 25.21 & 31.71 \\
        SDXL-turbo & Yes & No  & 21.24 & 31.54 \\
        \midrule
        \multicolumn{5}{c}{\textbf{COCO 2017 ($512\times512$)}} \\
        \midrule
        SDXL-turbo  & No  & --  & 32.08 & 31.67 \\
        SDXL-turbo  & Yes & No  & 27.69 & 31.52 \\
        \midrule
        \multicolumn{5}{c}{\textbf{COCO 2014 ($1024\times1024$)}} \\
        \midrule
        SDXL-Lightning & No  & --  & 22.53 & 31.25 \\
        SDXL-Lightning & Yes & Yes & 18.21 & 29.55 \\
        SDXL-Lightning & Yes & No  & 18.55 & 29.49 \\
        Hyper SDXL     & No  & --  & 24.97 & 31.50 \\
        Hyper SDXL     & Yes & Yes & 21.95 & 30.20 \\
        Hyper SDXL     & Yes & No  & 20.85 & 30.04 \\
        DMD2           & No  & --  & 19.35 & 31.75 \\
        DMD2           & Yes & Yes & 16.05 & 30.87 \\
        DMD2           & Yes & No  & 15.84 & 30.81 \\
        Flash SDXL     & No  & --  & 20.42 & 31.20 \\
        Flash SDXL     & Yes & Yes & 19.10 & 29.49 \\
        Flash SDXL     & Yes & No  & 18.95 & 29.52 \\
        \midrule
        \multicolumn{5}{c}{\textbf{COCO 2017 ($1024\times1024$)}} \\
        \midrule
        SDXL-Lightning & No  & --  & 28.99 & 31.21 \\
        SDXL-Lightning & Yes & Yes & 24.46 & 29.53 \\
        SDXL-Lightning & Yes & No  & 24.72 & 29.46 \\
        Hyper SDXL     & No  & --  & 31.79 & 31.46 \\
        Hyper SDXL     & Yes & Yes & 28.12 & 30.08 \\
        Hyper SDXL     & Yes & No  & 27.10 & 30.06 \\
        DMD2           & No  & --  & 25.36 & 31.77 \\
        DMD2           & Yes & Yes & 22.24 & 30.81 \\
        DMD2           & Yes & No  & 22.57 & 30.81 \\
        Flash SDXL     & No  & --  & 27.09 & 31.24 \\
        Flash SDXL     & Yes & Yes & 25.56 & 29.49 \\
        Flash SDXL     & Yes & No  & 25.62 & 29.49 \\
        \bottomrule
    \end{tabular}
\end{table}

\section{Conclusion}

We introduced EMPURPLE, a training-free method that improves the FID of distilled diffusion models by reducing the train--test distribution mismatch of intermediate noisy features. We connected this mismatch to a PAC-style generalization argument: forcing early detailed predictions increases the effective complexity of the target and can push intermediate states out of distribution. Across multiple distilled samplers, EMPURPLE consistently improves FID, but slightly decrease CLIP score. Meanwhile, in the future, researchers can try to train a distillation model starting in a special timestep.

\section{Further Discussion}

I want to briefly connects our PAC-style view to a classic religious debate. ``Probatio diabolica'' argues that: finding a demon can prove the existence of a demon, but failing to find a demon in daily life is not evidence of the nonexistence of the demon. Analogously, a law like $F=ma$ can fit all daily observations yet fail in an unseen corner case. In PAC terms, many hypotheses can explain the observed data; the question is why we should trust a particular one to generalize. The usual answer is simplicity through constraint. By restricting the hypothesis class (e.g., discouraging overly complex functions), we trade expressiveness for robustness. And that also is the reason why the discussion aiming at abstract problems like Probatio diabolica can relate to the classic mechanism.

A similar, romantic accident appears in the song \emph{Empurple}, which uses a specific purple (\#664f8c) to create a melody (\#66 4 f 8 c) via the drum, and then adds high-frequency music to enhance it. Our method similarly reuses blurred color blocks and adds high-frequency detail later to generate a detailed image. 

\clearpage
%
%
\bibliographystyle{splncs04}
\bibliography{main}

@String(AAAI  = {AAAI})

@article{song2020score,
  title={Score-based generative modeling through stochastic differential equations},
  author={Song, Yang and Sohl-Dickstein, Jascha and Kingma, Diederik P and Kumar, Abhishek and Ermon, Stefano and Poole, Ben},
  journal={arXiv preprint arXiv:2011.13456},
  year={2020}
}

@article{karras2022elucidating,
  title={Elucidating the design space of diffusion-based generative models},
  author={Karras, Tero and Aittala, Miika and Aila, Timo and Laine, Samuli},
  journal={Advances in neural information processing systems},
  volume={35},
  pages={26565--26577},
  year={2022}
}

@article{lu2022dpm,
  title={Dpm-solver: A fast ode solver for diffusion probabilistic model sampling in around 10 steps},
  author={Lu, Cheng and Zhou, Yuhao and Bao, Fan and Chen, Jianfei and Li, Chongxuan and Zhu, Jun},
  journal={Advances in neural information processing systems},
  volume={35},
  pages={5775--5787},
  year={2022}
}

@article{lu2025dpm,
  title={Dpm-solver++: Fast solver for guided sampling of diffusion probabilistic models},
  author={Lu, Cheng and Zhou, Yuhao and Bao, Fan and Chen, Jianfei and Li, Chongxuan and Zhu, Jun},
  journal={Machine Intelligence Research},
  pages={1--22},
  year={2025},
  publisher={Springer}
}

@article{zhao2023unipc,
  title={Unipc: A unified predictor-corrector framework for fast sampling of diffusion models},
  author={Zhao, Wenliang and Bai, Lujia and Rao, Yongming and Zhou, Jie and Lu, Jiwen},
  journal={Advances in Neural Information Processing Systems},
  volume={36},
  pages={49842--49869},
  year={2023}
}

@article{lin2024sdxllightning,
  title={Sdxl-lightning: Progressive adversarial diffusion distillation},
  author={Lin, Shanchuan and Wang, Anran and Yang, Xiao},
  journal={arXiv preprint arXiv:2402.13929},
  year={2024}
}

@article{luo2311lcm,
  title={Lcm-lora: A universal stable-diffusion acceleration module. arXiv 2023},
  author={Luo, S and Tan, Y and Patil, S and Gu, D and Von Platen, P and Passos, A and Huang, L and Li, J and Zhao, H},
  journal={arXiv preprint arXiv:2311.05556}
}

@inproceedings{sauer2024adversarial,
  title={Adversarial diffusion distillation},
  author={Sauer, Axel and Lorenz, Dominik and Blattmann, Andreas and Rombach, Robin},
  booktitle={European Conference on Computer Vision},
  pages={87--103},
  year={2024},
  organization={Springer}
}

@article{song2023consistency,
  title={Consistency models},
  author={Song, Yang and Dhariwal, Prafulla and Chen, Mark and Sutskever, Ilya},
  year={2023}
}

@inproceedings{yin2024dmd,
  title={One-step diffusion with distribution matching distillation},
  author={Yin, Tianwei and Gharbi, Micha{\"e}l and Zhang, Richard and Shechtman, Eli and Durand, Fredo and Freeman, William T and Park, Taesung},
  booktitle={Proceedings of the IEEE/CVF conference on computer vision and pattern recognition},
  pages={6613--6623},
  year={2024}
}

@article{yin2024improveddmd,
  title={Improved distribution matching distillation for fast image synthesis},
  author={Yin, Tianwei and Gharbi, Micha{\"e}l and Park, Taesung and Zhang, Richard and Shechtman, Eli and Durand, Fredo and Freeman, Bill},
  journal={Advances in neural information processing systems},
  volume={37},
  pages={47455--47487},
  year={2024}
}

@article{ho2022classifier,
  title={Classifier-free diffusion guidance},
  author={Ho, Jonathan and Salimans, Tim},
  journal={arXiv preprint arXiv:2207.12598},
  year={2022}
}

@inproceedings{zhou2024amed,
  title={Fast ode-based sampling for diffusion models in around 5 steps},
  author={Zhou, Zhenyu and Chen, Defang and Wang, Can and Chen, Chun},
  booktitle={Proceedings of the IEEE/CVF Conference on Computer Vision and Pattern Recognition},
  pages={7777--7786},
  year={2024}
}

@inproceedings{chadebec2025flash,
  title={Flash diffusion: Accelerating any conditional diffusion model for few steps image generation},
  author={Chadebec, Clement and Tasar, Onur and Benaroche, Eyal and Aubin, Benjamin},
  booktitle={Proceedings of the AAAI Conference on Artificial Intelligence},
  volume={39},
  number={15},
  pages={15686--15695},
  year={2025}
}

@inproceedings{lin2014coco,
  title={Microsoft coco: Common objects in context},
  author={Lin, Tsung-Yi and Maire, Michael and Belongie, Serge and Hays, James and Perona, Pietro and Ramanan, Deva and Doll{\'a}r, Piotr and Zitnick, C Lawrence},
  booktitle={European conference on computer vision},
  pages={740--755},
  year={2014},
  organization={Springer}
}

@article{liu2022rectified,
  title={Rectified flow: A marginal preserving approach to optimal transport},
  author={Liu, Qiang},
  journal={arXiv preprint arXiv:2209.14577},
  year={2022}
}

@article{bertrand2025closed,
  title={On the Closed-Form of Flow Matching: Generalization Does Not Arise from Target Stochasticity},
  author={Bertrand, Quentin and Gagneux, Anne and Massias, Mathurin and Emonet, R{\'e}mi},
  journal={arXiv preprint arXiv:2506.03719},
  year={2025}
}

@misc{li2026fschedule,
      title={F-scheduler: illuminating the free-lunch design space for fast sampling of diffusion models}, 
      author={Zilai Li and Lujia Bai},
      year={2026},
      eprint={2510.02390},
      archivePrefix={arXiv},
      primaryClass={cs.GR},
      url={https://arxiv.org/abs/2510.02390}, 
}

@article{yu2021frechet,
  title={Frechet inception distance (fid) for evaluating gans},
  author={Yu, Yanchun and Zhang, Weibin and Deng, Yun},
  journal={China University of Mining Technology Beijing Graduate School},
  volume={3},
  number={11},
  year={2021}
}

@article{heusel2017gans,
  title={Gans trained by a two time-scale update rule converge to a local nash equilibrium},
  author={Heusel, Martin and Ramsauer, Hubert and Unterthiner, Thomas and Nessler, Bernhard and Hochreiter, Sepp},
  journal={Advances in neural information processing systems},
  volume={30},
  year={2017}
}

@article{goodfellow2020generative,
  title={Generative adversarial networks},
  author={Goodfellow, Ian and Pouget-Abadie, Jean and Mirza, Mehdi and Xu, Bing and Warde-Farley, David and Ozair, Sherjil and Courville, Aaron and Bengio, Yoshua},
  journal={Communications of the ACM},
  volume={63},
  number={11},
  pages={139--144},
  year={2020},
  publisher={ACM New York, NY, USA}
}

@article{saatci2017bayesian,
  title={Bayesian gan},
  author={Saatci, Yunus and Wilson, Andrew G},
  journal={Advances in neural information processing systems},
  volume={30},
  year={2017}
}

@incollection{ketkar2017stochastic,
  title={Stochastic gradient descent},
  author={Ketkar, Nikhil},
  booktitle={Deep learning with Python: A hands-on introduction},
  pages={113--132},
  year={2017},
  publisher={Springer}
}

@article{hoeffding1963probability,
  title={Probability inequalities for sums of bounded random variables},
  author={Hoeffding, Wassily},
  journal={Journal of the American statistical association},
  volume={58},
  number={301},
  pages={13--30},
  year={1963},
  publisher={Taylor \& Francis}
}

@article{ruder2016overview,
  title={An overview of gradient descent optimization algorithms},
  author={Ruder, Sebastian},
  journal={arXiv preprint arXiv:1609.04747},
  year={2016}
}

@inproceedings{mokady2023null,
  title={Null-text inversion for editing real images using guided diffusion models},
  author={Mokady, Ron and Hertz, Amir and Aberman, Kfir and Pritch, Yael and Cohen-Or, Daniel},
  booktitle={Proceedings of the IEEE/CVF conference on computer vision and pattern recognition},
  pages={6038--6047},
  year={2023}
}

@inproceedings{sohl2015deep,
  title={Deep unsupervised learning using nonequilibrium thermodynamics},
  author={Sohl-Dickstein, Jascha and Weiss, Eric and Maheswaranathan, Niru and Ganguli, Surya},
  booktitle={International conference on machine learning},
  pages={2256--2265},
  year={2015},
  organization={pmlr}
}

@book{deco1996information,
  title={An information-theoretic approach to neural computing},
  author={Deco, Gustavo and Obradovic, Dragan},
  year={1996},
  publisher={Springer Science \& Business Media}
}

@article{pearson1901liii,
  title={LIII. On lines and planes of closest fit to systems of points in space},
  author={Pearson, Karl},
  journal={The London, Edinburgh, and Dublin philosophical magazine and journal of science},
  volume={2},
  number={11},
  pages={559--572},
  year={1901},
  publisher={Taylor \& Francis}
}

@inproceedings{roy2007effective,
  title={The effective rank: A measure of effective dimensionality},
  author={Roy, Olivier and Vetterli, Martin},
  booktitle={2007 15th European signal processing conference},
  pages={606--610},
  year={2007},
  organization={IEEE}
}

@book{cover1999elements,
  title={Elements of information theory},
  author={Cover, Thomas M},
  year={1999},
  publisher={John Wiley \& Sons}
}

@article{ren2024hyper,
  title={Hyper-sd: Trajectory segmented consistency model for efficient image synthesis},
  author={Ren, Yuxi and Xia, Xin and Lu, Yanzuo and Zhang, Jiacheng and Wu, Jie and Xie, Pan and Wang, Xing and Xiao, Xuefeng},
  journal={Advances in neural information processing systems},
  volume={37},
  pages={117340--117362},
  year={2024}
}

@inproceedings{hessel2021clipscore,
  title={Clipscore: A reference-free evaluation metric for image captioning},
  author={Hessel, Jack and Holtzman, Ari and Forbes, Maxwell and Le Bras, Ronan and Choi, Yejin},
  booktitle={Proceedings of the 2021 conference on empirical methods in natural language processing},
  pages={7514--7528},
  year={2021}
}

\clearpage
\appendix
\section{Appendix: Variance Bias Trade-off For SGD}
\label{app:vb-tradeoff}

For a neural network with a given optimization method and the hyperparameters, the function it can express after the training with $i$ times iteration constitutes the hypothesis space $\mathcal{H}$, and we can define a special function it can express as a hypothesis $h$ by providing a particular training dataset $d\sim D^m$.  When given a special training dataset $d$ and a input $X$, for a deterministic neural network, the $i$th gradient descent (GD) \cite{ruder2016overview} optimizaiton would provide a fixed value as output, while the possible output $Y$ with given $X$ in $i$th stochastic gradient descent (SGD) \cite{ketkar2017stochastic} optimization is a random variables, and typically, we can use the log likelihood to measure the performance.  However, since we don't know the probability distribution of the output, we deem th $i$th iteration provide us a series of functions with different possibility density, and consider the loss in \cref{eq:new metrics appendix}, which $\mathcal{F}$ denote all of the possible function the model can express in the $i$th iteration on this special SGD optimization with the given training dataset $d$.  
\begin{equation}
\mathcal{L}(f^d(X),Y)=\mathbb{E}_{f\sim \mathcal{F}}[(f(X)-Y)^2]
  \label{eq:new metrics appendix}
\end{equation}

First, we expend the loss via bias-variance trade-off.
\begin{align}
\nonumber 
\mathcal{L}(f^d(X),Y)=&\mathbb{E}_{f\sim \mathcal{F}}[(f(X)-Y)^2]\\ \nonumber 
=& \mathbb{E}_{f\sim \mathcal{F}}[(f(X)-\hat{f}^d(X)+\hat{f}^d(X)-Y)^2] \\ \nonumber
=& \mathbb{E}_{f\sim \mathcal{F}}[(f(X)-\hat{f}^d(X))^2+2(f(X)-\hat{f}^d(X))(\hat{f}^d(X)-Y)\\ \nonumber &+(\hat{f}^d(X)-Y)^2] \\ \nonumber
=&\mathbb{E}_{f\sim \mathcal{F}}[(f(X)-\hat{f}^d(X))^2]+2(\hat{f}^d(X)-Y)\mathbb{E}_{f\sim \mathcal{F}}[f(x)-\hat{f}^d(X)] \\ \nonumber &+\mathbb{E}_{f\sim \mathcal{F}}[(\hat{f}^d(X)-Y)^2] \\ \nonumber
=& \mathbb{E}_{f\sim \mathcal{F}}[(f(X)-\hat{f}^d(X))^2] + \mathbb{E}_{f\sim \mathcal{F}}[(\hat{f}^d(X)-Y)^2]\\ 
=&  \mathbb{E}_{f\sim \mathcal{F}}[(f(X)-\hat{f}^d(X))^2] + (\hat{f}^d(X)-Y)^2
\label{loss-decompose}
\end{align}
, where $\hat{f}^d(x)=E_{f\sim\mathcal{F}}[f(x)]$

Then, for the input of neural network, a noisy image $X$, and the target output, a predict clean image $T(X)$ given by a neural network in timestep $t$ whose multi-step inference result has a better FID performance, we consider the expected loss $\mathbb{E}_{X\sim \mathcal{X},Y=T(X)}[\mathcal{L}(f^d(X),Y)]$ and empirical loss $\mathbb{E}_{X\sim d,Y=T(X)}[\mathcal{L}(f^d(X),Y)]$. 
\begin{align}
   \nonumber &\mathbb{E}_{X\sim \mathcal{X},Y=T(X)}[\mathcal{L}(f^d(X),Y)]\\ \nonumber =& \mathbb{E}_{X\sim \mathcal{X},Y=T(X)}[(\hat{f}^d(X)-Y)^2] +\mathbb{E}_{X\sim \mathcal{X},f\sim \mathcal{F}}[(f(X)-\hat{f}^d(X))^2] \\ \nonumber
   =& \mathbb{E}_{X\sim \mathcal{X},Y=T(X)}[(\hat{f}^d(X)-f'(x)+f'(X)-Y)^2] +\mathbb{E}_{X\sim \mathcal{X},f\sim \mathcal{F}}[(f(X)-\hat{f}^d(X))^2] \\ \nonumber
   =& \mathbb{E}_{X\sim \mathcal{X},Y=T(X)}[(\hat{f}^d(X)-f'(X))^2 + 2(\hat{f}^d(X)-f'(X))( f'(X)-Y)\\ \nonumber +&(f'(X)-Y)^2] +\mathbb{E}_{X\sim \mathcal{X},f\sim \mathcal{F}}[(f(X)-\hat{f}^d(X))^2] \\ \nonumber
   = &\mathbb{E}_{X\sim \mathcal{X}}[(\hat{f}^d(X))^2-2\hat{f}^d(X)f'(X)+(f'(X))^2]\\ \nonumber +& \mathbb{E}_{X\sim \mathcal{X},Y=T(X)}[2\hat{f}^d(X)f'(X)-2\hat{f}^d(X)Y-2(f'(X))^2+2f'(X)Y\\ \nonumber+&(f'(X))^2-2f'(X)Y+Y^2]+\mathbb{E}_{X\sim \mathcal{X},f\sim \mathcal{F}}[(f(X)-\hat{f}^d(X))^2] \\ \nonumber =&\mathbb{E}_{X\sim \mathcal{X}}[(\hat{f}^d(X))^2] - (f'(X))^2 \\ \nonumber +&\mathbb{E}_{X\sim \mathcal{X},Y=T(X)}[2\hat{f}^d(X)f'(X)-2\hat{f}^d(X)Y-2(f'(X))^2+2f'(X)Y\\ \nonumber+&(f'(X))^2-2f'(X)Y+Y^2]+\mathbb{E}_{X\sim \mathcal{X},f\sim \mathcal{F}}[(f(X)-\hat{f}^d(X))^2] \\ \nonumber =&\mathbb{E}_{X\sim \mathcal{X},Y=T(X)}[(f'(X)-Y)^2] + \mathbb{E}_{X\sim \mathcal{X},f\sim \mathcal{F}}[(f(X)-\hat{f}^d(X))^2]\\ \nonumber+&\mathbb{E}_{X\sim \mathcal{X}}[(\hat{f}^d(X))^2] - (f'(X))^2 \\ \nonumber
   =& \mathbb{E}_{X\sim \mathcal{X},Y=T(X)}[(f'(X)-Y)^2] + \mathbb{E}_{x \sim\mathcal{X},f\sim \mathcal{F}}[(f(X))^2]-(f'(X))^2 \\ \nonumber
   =& \mathbb{E}_{X\sim \mathcal{X},Y=T(X)}[(f'(X)-Y)^2] +  \mathbb{E}_{x \sim\mathcal{X},f\sim \mathcal{F}}[(f(X))^2]- \mathbb{E}_{x \sim\mathcal{X},f\sim \mathcal{F}}[f(X)]^2  \\ \nonumber
   =& \mathbb{E}_{X\sim \mathcal{X},Y=T(X)}[(f'(X)-Y)^2] + \mathbb{E}_{X\sim \mathcal{X},f\sim \mathcal{F}}[(f(X)-f'(X))^2]
\end{align}
, in which $f'(x)=E_{X\sim \mathcal{X}}[\hat{f}^d(x)]$, and when the hyperparameters of the neural network as well as the optimizer are given, the iteration of the optimization is $i$, then it's a fixed value.

The first equation is obtained by replacing the $\mathcal{L}$ with the \cref{loss-decompose}.  For the sixth equation, we get by first calculating the expectation of $X$ on the right part of the fifth equation. For the seventh equation, we use the definition of the variance $\mathbb{E}_X[(X-E(X))^2]=\mathbb{E}_X[X^2]-(\mathbb{E}_X[X])^2$ to expand the second term in the right equation.  And similar to the ninth equation.

Similarly, for the empirical loss, we can write as $\mathbb{E}_{X\sim S,Y=T(X)}[\mathcal{L}(f^d(X),Y)]$, and it has:
\begin{align}
    &\nonumber\mathbb{E}_{X\sim S,Y=T(X)}[\mathcal{L}(f^d(X),Y)]  \\ =& \mathbb{E}_{X\sim S,Y=T(X)}[(f''(X)-Y)^2] + \mathbb{E}_{X\sim S,f\sim \mathcal{F}}[(f(X)-f''(X))^2]
\end{align}
, in which $f''(x)=E_{X\sim S}[\hat{f}^d(x)]$

\section{Appendix: PAC bound based on mutual information}
\label{app:pac-b-mi}

Then, for a function with $|\Gamma|$ possible output and $|\Omega|$ possible input, the upper bound of the $|\mathcal{H}|$ can be write as:
\begin{equation}
    |\mathcal{H}|\leq |\Gamma|^{|\Omega|}
\end{equation}

And the volume of $Y$ in the typical set is $2^{H(Y)}$.  Meanwhile, for the given input $X$, we also have a conditional typical set, which denotes the typical $Y$ when $X$ is given. In computer science, we don't handle infinite precision input, and the amount of possible output $Y$ is  $c\cdot2^{H(Y)}$. and $2^{I(X;Z)}=\frac{2^{H(Y)}}{2^{H(Y|X)}}$, which denote the volume of meaningful input.  Meanwhile, the volume of meaningful input is equal to the volume of the meaningful output, since $2^{I(X;Z)}=\frac{2^{H(X)}}{2^{H(X|Y)}}$.

Hence the test-training loss can be wroten as:
\begin{align}
    \nonumber\Delta(s) &= \mathbb{E}_{X\sim \mathcal{X},Y=T(X)}[(f'(X)-Y)^2]  - \mathbb{E}_{X\sim S,Y=T(X)}[(f''(X)-Y)^2]\\ \nonumber &+ \mathbb{E}_{X\sim \mathcal{X},f\sim \mathcal{F}}[(f(X)-f'(X))^2] - \mathbb{E}_{X\sim S,f\sim \mathcal{F}}[(f(X)-f''(X))^2]\\\nonumber &\leq\sqrt{\frac{\log{|\mathcal{H}|}+\log{\frac{2}{\sigma}}}{2o}} \\ \nonumber
    &\leq \sqrt{\frac{\log{(2^{I(X;Y)})^{2^{I(X;Y)}}+\log{\frac{2}{\sigma}}}}{2o}}\\
    &\leq \sqrt{\frac{{I(X;Y)}2^{I(X;Y)}+\log{\frac{2}{\sigma}}}{2o}}
\end{align}

\end{document}